\documentclass[letterpaper, 10 pt, conference]{format/ieeeconf}  

\IEEEoverridecommandlockouts                              

\usepackage{dsfont}

\usepackage{amsmath}

\usepackage{graphics} 
\usepackage{graphicx}
\usepackage{amsmath} 
\usepackage{amssymb}  
\usepackage{todonotes}

\usepackage{amsfonts}
\usepackage{mathtools}
\usepackage[linesnumbered,algoruled,boxed,vlined, noend]{algorithm2e}
\usepackage{amsthm}
\usepackage{comment}
\usepackage{cite}
\usepackage{wrapfig}
\let\labelindent\relax
\usepackage{enumitem}
\usepackage{overpic}
\usepackage{xspace}
\usepackage{multirow}
\usepackage{authblk}

\DeclareMathAlphabet{\mathcal}{OMS}{cmsy}{m}{n} 

\newtheorem{problem}{Problem}[section]

\theoremstyle{definition}

\theoremstyle{remark}

\DeclarePairedDelimiterX{\norm}[1]{\lVert}{\rVert}{#1}

{\end{list}}

\usepackage{soul} 
\usepackage[a-2b,mathxmp]{pdfx}[2018/12/22]

\def\prob{MoTaR\xspace}
\def\model{StabilNet\xspace}
\def\astar{A*\xspace}
\def\ours{ORLA*\xspace}

\newif\ifarxiv
\arxivfalse

\font\titlefont=ptmb at 15.95pt
\title{\titlefont
ORLA*: Mobile Manipulator-Based Object Rearrangement with Lazy A* }
\author{Kai Gao$^{*,1}$\quad Zhaxizhuoma$^{*,2}$\quad Yan Ding$^{3}$\quad Shiqi Zhang$^{3}$\quad Jingjin Yu$^{1}$
\thanks{$^*$These authors made equal contributions to the study.}
}
\begin{document}

\maketitle


\begin{abstract}
Effectively performing object rearrangement is an essential skill for mobile manipulators, e.g., setting up a dinner table. A key challenge in such problems is deciding an appropriate ordering to effectively untangle object-object dependencies while considering the necessary motions for realizing the manipulations (e.g., pick and place). To our knowledge, computing time-optimal multi-object rearrangement solutions for mobile manipulators remains a largely untapped research direction. In this work, we propose \ours, which leverages delayed/lazy evaluation in searching for a high-quality object pick-n-place sequence that considers both end-effector and mobile robot base travel. \ours readily handles multi-layered rearrangement tasks powered by learning-based stability predictions. Employing an optimal solver for finding temporary locations for displacing objects, \ours can achieve global optimality. Through extensive simulation and ablation study, we confirm the effectiveness of \ours delivering quality solutions for challenging rearrangement instances. Supplementary materials are available at:
\href{https://gaokai15.github.io/ORLA-Star/}
{\textcolor{blue}{\texttt{gaokai15.github.io/ORLA-Star/}}}
\end{abstract}

\section{Introduction}\label{sec:intro}
\begin{figure}[t]
\vspace{2mm}
    \centering
    \includegraphics[width=1\columnwidth]{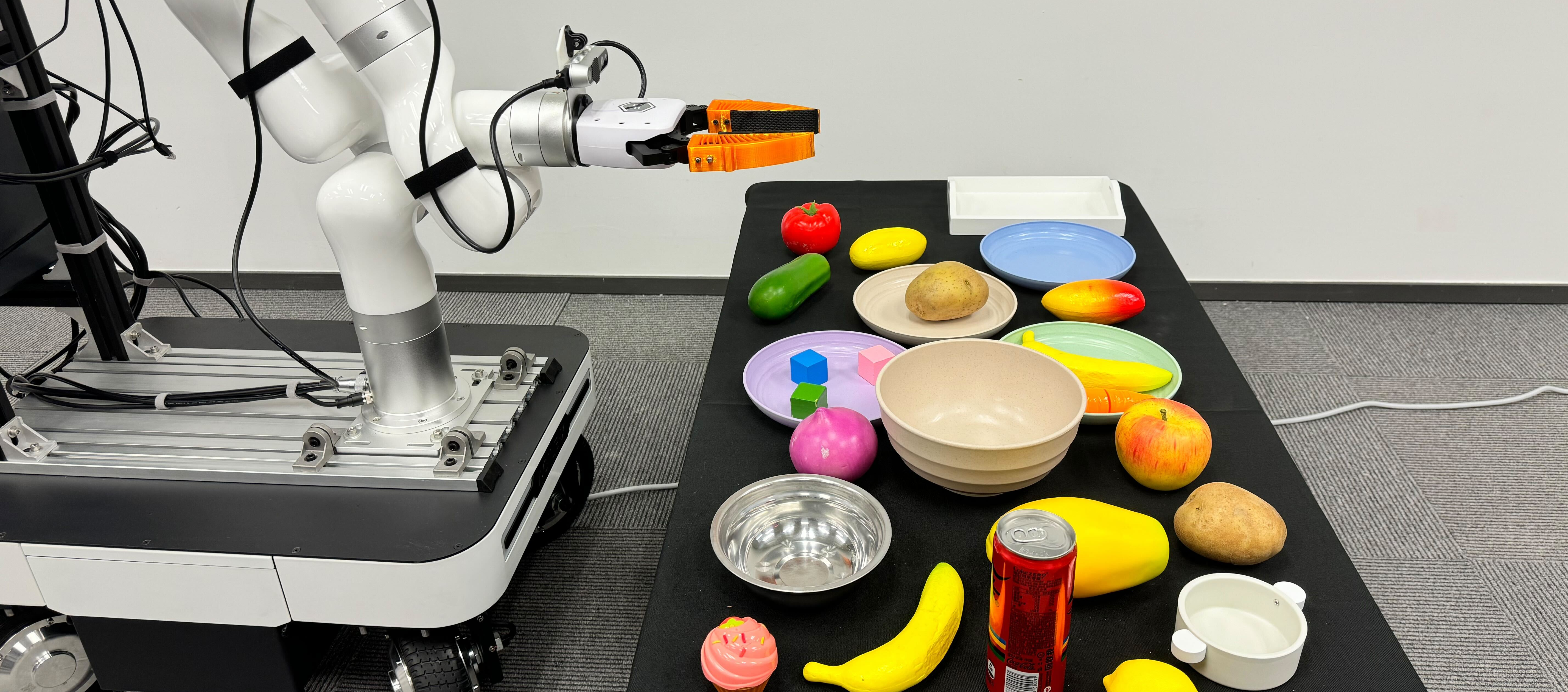}
    \caption{An example Mobile Robot Tabletop Rearrangement (\prob) setup.}
    \label{fig:intro}
\end{figure}

Prominent robotics startups have promised to deliver mobile robotic solutions for everyday tasks, e.g., having a humanoid robot skillfully manipulating objects and/or transporting them. 
To realize that lofty promise, mobile manipulators must solve object rearrangement tasks, e.g., tidying up a large work area or setting/cleaning a dinner table, with great efficiency and simultaneously producing natural plans - consider, for example, performing rearrangement for the setup illustrated in Fig.~\ref{fig:intro} to reach a more ordered target arrangement. 
Tackling such practical settings demands algorithms capable of computing highly efficient rearrangement plans where object manipulation sequence planning must be tightly coupled with planning the movement of the mobile manipulator's base. Furthermore, a comprehensive solution must consider complex rearrangement, e.g., placing fruits in bowls or piling books. In such cases, \emph{stability} must be carefully considered. For example, placing an apple on a plate is generally stable but not the opposite. 

Somewhat surprisingly and unfortunately, even for fixed manipulators with direct access to the entire workspace, apparently ``simple'' tabletop rearrangement settings prove to be computationally intractable to optimally solve \cite{han2018complexity,gao2023minimizing}, dimming hopes for finding extremely scalable algorithmic solutions for such problems. 
The challenges arise from deciding on a high-quality sequence of manipulation actions (e.g., pick and place) to avoid redundant actions, which require carefully untangling intricate dependencies between the objects.    
Nevertheless, effective solvers have been proposed for practical-sized tabletop rearrangement problems in dense settings \cite{gao2023minimizing,huang2019large,labbe2020monte} for manipulators with fixed bases, leveraging a careful fusion of combinatorial reasoning and systematic search.   
Solutions addressing stability issues in multi-layer rearrangement problems have also been proposed \cite{xu2023optimal}. 

This work develops effective solutions for mobile manipulators over a larger workspace and jointly considers the arrangement's stability. 
To that end, we propose \ours: \emph{Object Rearrangement with Lazy A*} for solving mobile manipulator-based rearrangement tasks. 
We carefully investigated factors impacting the optimality of a rearrangement plan for mobile manipulators and provided insightful structural understandings for the same. Among these, a particularly interesting one is that the mobile base travels on $S^1$ (i.e., a cycle), leading to intricate interactions with other factors in the optimization task. 
%
\ours designs a suitable cost function that integrates the multiple costs and employs the idea of lazy buffer allocation (buffers are temporary locations for objects that cannot be placed at their goals) into the \astar framework to search for rearrangement plans minimizing the cost function.
To accomplish this, we redefine the $f,g,h$ values in \astar when some states in the search tree are non-deterministic due to the delayed buffer computation.
With optimal buffer computation, \ours returns globally optimal solutions.
A thorough feasibility and optimality study backs our buffer allocation strategies. To estimate the feasibility of a buffer pose, especially when we want to temporarily place an object on top of others, we propose a learning model, \model, to estimate the stability of the placing pose. 

\section{Related Works}\label{sec:related}
{\bf Tabletop Rearrangement} In tabletop rearrangement tasks, the primary challenge lies in planning a long sequence of actions in a cluttered environment.
Such rearrangement planning can be broadly divided into three primary categories: prehensile\cite{zeng2021transporter,gao2023minimizing,han2018complexity,zhang2022visually,gao2022fast,labbe2020monte,ding2023task, xu2023optimal}, non-prehensile\cite{yuan2019end,vieira2022persistent,huang2021dipn,song2020multi,huang2021visual}, and a combination of the two\cite{tang2023selective}.
Compared with non-prehensile operations (e.g., pushing and poking), prehensile manipulations, while demanding more precise grasping poses\cite{zeng2021transporter} prior to picking, offer the advantage of placing objects with higher accuracy in desired positions and facilitate planning over longer horizons\cite{gao2022toward}.
In this domain, commonly employed cost functions encompass metrics like the total count of actions\cite{gao2022fast,gao2023minimizing,xu2023optimal}, execution duration\cite{gao2022toward}, and end-effector travel\cite{han2018complexity,song2020multi}, among others\cite{gao2023effectively}.
In this paper, we manipulate objects with overhand pick-n-places. 
Besides end-effector traveling costs, we also propose a novel cost function considering the traveling cost of a mobile robot.

{\bf Buffer Allocation} In the realm of rearrangement problems, there are instances where specific objects cannot be moved directly to their intended goal poses. 
These scenarios compel the temporary movement of objects to collision-free poses. 
To streamline the rearrangement planning, several rearrangement methodologies exploit external free spaces as buffer zones\cite{gao2023minimizing,xu2023optimal}. 
One notable concept is the \emph{running buffer size}\cite{gao2023minimizing}, which quantifies the requisite size of this external buffer zone. 
In situations devoid of external space for relocation, past research either pre-identifies potential buffer candidates\cite{wang2021uniform,cheong2020relocate} or segments the rearrangement tasks into sequential subproblems\cite{krontiris2016efficiently,wang2022lazy}.
TRLB\cite{gao2022fast}, aiming for an optimized buffer selection, prioritizes task sequencing and subsequently employs the task plan to dictate buffer allocation. 
However, TRLB doesn't factor in travel costs. 
Contrarily, in our study, we incorporate lazy buffer allocation within the \astar search and prioritize buffer poses based on various cost function optimizations.


{\bf Manipulation Stability} 
Structural stability is pivotal in robot manipulation challenges. 
Wan et al.\cite{wan2018assembly} assess the stability of Tetris blocks by scrutinizing their supporting boundaries. 
For truss structures, finite element methods have been employed to assess stability of intermediate stages\cite{mcevoy2014assembly,garrett2020scalable}. 
Utilizing deep learning, Noseworth et. al.\cite{noseworthy2021active} introduce a Graph Neural Network model dedicated to evaluating the stability of stacks of cuboid objects. 
However, these methodologies often come with shape constraints, requiring objects to be in forms such as cuboid blocks or truss structures. 
For objects of more general shapes, recent research\cite{xu2023optimal,lee2023object} leverages stability checkers grounded in physics simulators, which are effective but tend to be computationally intensive. 
In contrast, our study introduces a deep-learning-based prediction model, \model, tailored for the rearrangement of objects with diverse shapes. 
\model offers a speed advantage over simulation-based checks and demonstrates robust generalization to previously unseen object categories.

\section{Problem Formulation}\label{sec:prob}
Consider a 2D tabletop workspace centered at the origin of the world coordinate with $z$ pointing up. 
A point $(x,y,z)$ is within the tabletop region $\mathcal W$ if $(x,y)$ are contained in the workspace and $z \geq 0$. 
The workspace has $n$ objects $\mathcal{O}$. 
The pose of a workspace object $o_i\in \mathcal O$ is represented as $(x, y, z, \theta)$.
An arrangement of $\mathcal O$ is feasible if all objects are contained in the tabletop region and are collision-free.
In this paper, we allow objects to be placed on top of others.
An object is \emph{graspable} at this arrangement if no other object is on top.
A goal pose is \emph{available} if both conditions below are true: (1). There is no obstacle blocking the pose; (2). If other objects have goal poses under the pose, these objects should have been at the goal poses.

On the table's edge, a mobile robot equipped with an arm moves objects from an initial arrangement $\mathcal A_I$ to a desired goal arrangement $\mathcal A_G$ with overhand pick and place actions.
Each action $a$ is represented by $(o_i, p_1, p_2)$, which moves object $o_i$ from current pose $p_1$ to a collision-free target pose $p_2$.
A rearrangement plan $\Pi=\{a_1, a_2, a_3, ...\}$ is a sequence of actions moving objects from $\mathcal A_I$ to $\mathcal A_G$.

We evaluate solution quality with a cost function $J(\Pi)$ (Eq.\ref{eq:cost}), which mimics the execution time of the plan. 
The first term is the total traveling cost, and the second is the manipulation cost associated with pick-n-places, which is linearly correlated with the number of actions.
\vspace{1mm}
\begin{align}\label{eq:cost}
    J(\Pi)&=dist(\Pi)+mani(\Pi),  \ 
    mani(\Pi)=C|\Pi|
\end{align}

Based on the description so far, we define the studied problem as follows.

\begin{problem}[Mobile Robot Tabletop Rearrangement (\prob)]
    Given a feasible initial arrangement $\mathcal A_I$ and feasible goal arrangement $\mathcal A_G$ of an object set $\mathcal O$, find the rearrangement plan $\Pi$ minimizing the cost $J(\Pi)$.
\end{problem}

\begin{figure}[h]
    \centering
    \includegraphics[width=\columnwidth]{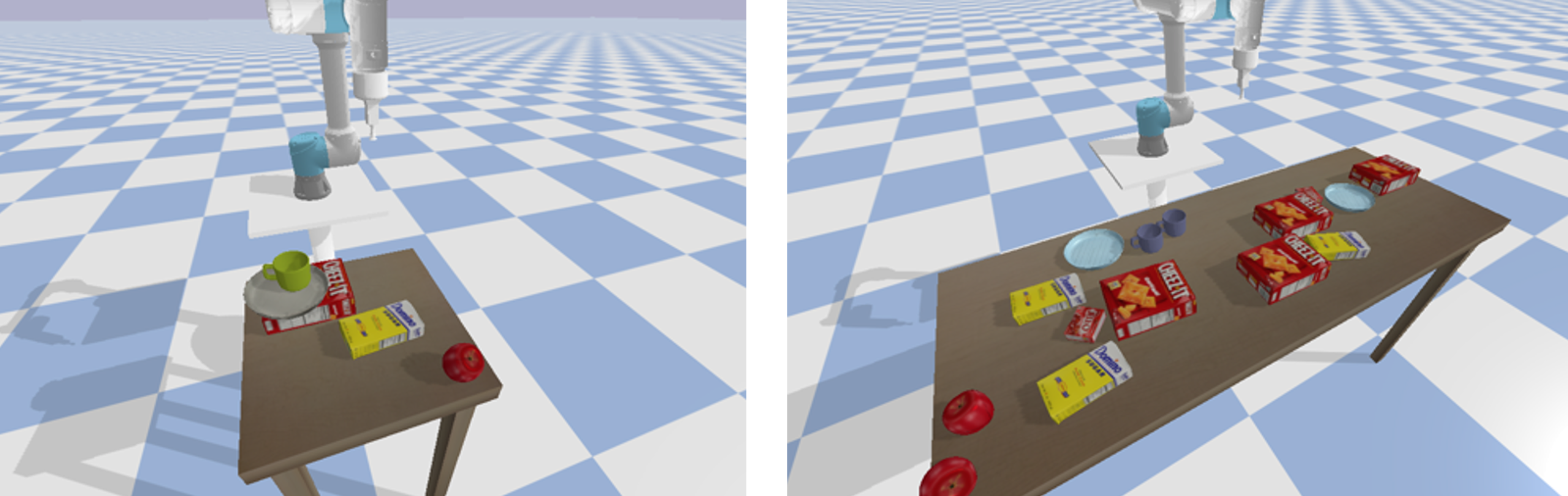}
    \caption{[Left] An example of the EE scenario, where the table is small and the robot can reach all poses from a fixed position. We count the traveling cost of the end-effector (EE) in the cost function. [Right] An example of the MB scenario, where the table is large and the robot can only reach a portion of tabletop poses from a fixed position. We count the traveling cost of the mobile base (MB) in the cost function.}
    \label{fig:scenarios}
\end{figure}

We study \prob under two scenarios.
On the one hand, for a small workspace ( Fig.~\ref{fig:scenarios}[Left]), where the mobile robot can reach all tabletop poses at a fixed base position.
We define $dist()$ as the Euclidean travel distance of the end effector (EE) in the x-y plane.
On the other hand, for a large table workspace( Fig.~\ref{fig:scenarios}[Right]), where the mobile base needs to travel around to reach the picking/placing poses, we define $dist()$ as the Euclidean distance that the mobile base (MB) travels.
As shown in Fig.~\ref{fig:workingExample}[Left], we assume the mobile base travels along the boundaries of the table. 
When the robot attempts to pick/place an object at pose $(x,y,z,\theta)$, it will move to the closest point on the track to $(x,y)$ before executing the pick/place.
In the remainder of this paper, we will refer to the scenarios as \emph{EE} and \emph{MB}, respectively.

\section{\ours: A* with Lazy Buffer Allocation}\label{sec:planner}
We describe \ours, a \emph{lazy} A*-based rearrangement planner that delays buffer computation, specially designed for mobile manipulator-based object rearrangement problems.
As a variant of A*, \ours always explores the state $s$ that minimizes the estimated cost $f(s)=g(s)+h(s)$. 
An action from $s$ to its neighbor moves an object to the goal pose or a buffer.
Specifically, actions from $s$ follow the rules below:
\begin{enumerate}[leftmargin=5mm]
    \item \textbf{R1.} If $o_i$ is graspable and its goal pose is also available at $s$, move $o_i$ to its goal.
    \item \textbf{R2}. If $o_i$ is graspable, its goal is unavailable, and it causes another object to violate R1, then move $o_i$ to a buffer.
\end{enumerate}
When \ours decides to place an object at a buffer, it does not allocate the buffer pose immediately. 
Instead, the buffer pose is decided after the object leaves the buffer.
In this way, lazy buffer allocation effectively computes high-quality solutions with a low number of actions\cite{gao2022fast}.
Under the A* framework, \ours searches for buffer poses with the minimum additional traveling cost.

\subsection{Deterministic and Nondeterministic States}
To enable lazy buffer allocation, \ours categorizes states into deterministic states (DS) and non-deterministic states (NDS).
Like traditional \astar, a DS represents a feasible object arrangement in the workspace.
Each object has a deterministic pose at this state.
NDS is a state where some object is at a buffer pose to be allocated.

\begin{figure}[ht]
    \centering
    \includegraphics[width=0.16\textwidth]{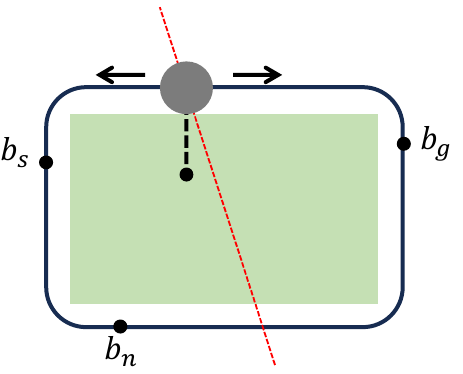}
    \hspace{6mm}
    \includegraphics[width=0.18\textwidth]{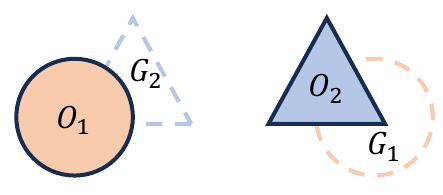}
    \caption{[Left] An example of MB scenario, where the robot (gray disc) travels along the green table following the black track along the table boundaries. 
    To pick/place an object on the table, the robot moves to the nearest position before the manipulation. 
    [Right] A working example where $o_1$ and $o_2$ block each other's goal pose. One must move to a buffer pose to finish the rearrangement.
    }
    \label{fig:workingExample}
\end{figure}

In the working example in Fig.~\ref{fig:workingExample}[Right], $o_1$ and $o_2$ blocks each other from goal poses.
The path in the A* search tree from the initial state $(p_1^I,p_2^I)$ to the goal state $(p_1^G,p_2^G)$ may be:
\vspace{-1mm}
\begin{small}
$$
S_1:(p_1^I,p_2^I)\to S_2:(B_1,p_2^I)\to S_3:(B_1,p_2^G)\to S_4: (p_2^G,p_2^G)
$$
\end{small}

In this rearrangement plan, the robot moves $o_1$ to buffer and then moves $o_2$ and $o_1$ to goal poses, respectively.
In this example, the initial and goal states $S_1$ and $S_4$ are DS, and both intermediate states $S_2$ and $S_3$ are NDSs since $o_1$ is at a non-deterministic buffer $B_1$ in these states.

\subsection{Cost Estimation}\label{sec:cost}
In \astar, each state $s$ in the search space is evaluated by $f(s)=g(s)+h(s)$, where $g(s)$ represents the cost from the initial state to $s$ and $h(s)$ represents the estimated cost from $s$ to the goal state.
\ours defines $g(s)$ and $h(s)$ for DS and define $f(s)$ for NDS.
For a DS $s_D$, $g(s_D)$ is represented by the actual cost measured by $\mathcal J(\cdot)$ from the initial state to $s_D$, which can be computed as follows:
$$
g(s_D) = g(s_D') + \mathcal J(s_D',s_D)
$$
where $s_D'$ is the last DS in this path, and $\mathcal J(s_D',s_D)$ is the cost of the path between $s_D'$ and $s_D$.
When there are NDSs between $s_D'$ and $s_D$, we compute buffer poses minimizing the cost.
The details are discussed in Sec.~\ref{sec:allocation}.
In $h(s_D)$ computation, we only count the transfer path and manipulation costs of one single pick-n-place for each object away from goal poses.
For example, in Fig.~\ref{fig:workingExample}, $h(S_1)=dist(p_1^s,p_1^g)+dist(p_2^s,p_2^g)+C*|\{o_1, o_2\}|$.

For an NDS $s_{N}$, we have
\begin{align}
\begin{split}
f(s_{N}) &= g(s_D') + \mathcal J(s_D',s_{N}) + h(s_{N})
\end{split}
\end{align}
Since the $g(x)$ computation of DSs and NDSs only relies on $g(\cdot)$ of $S_D'$, rather than any NDS, we do not compute $g(s_{N})$ explicitly. 
Instead, we directly compute the lower bound of the actual cost as $f(s_{N})$.
The general idea of $f(s_{N})$ computation is presented in Algo.\ref{alg:fx}.
In Line 2, we add all manipulation costs in $\mathcal J(s_D',s_N)$ and $h(s_N)$, which are defined in the same spirit as those of $s_D$.
In Lines 3-4, we add the traveling cost along deterministic poses between $s_D'$ and $s_N$, 
which is a lower bound of the actual traveling cost as it assumes buffer poses do not induce additional costs.
In Lines 5-8, we add traveling cost to $h(s_N)$.
If the object is at a buffer, we add traveling distance based on Algo.~\ref{alg:refine}, which we will mention more details later.
If the object is at a deterministic pose, we add the traveling cost of the transfer path between the current pose and its goal pose.
In our implementation, we store buffer information in each NDS to avoid repeated computations in Algo.~\ref{alg:fx}.

\begin{algorithm}
\begin{small}
    \SetKwInOut{Input}{Input}
    \SetKwInOut{Output}{Output}
    \SetKwComment{Comment}{\% }{}
    \caption{ $f(s_{N})$ Computation}
		\label{alg:fx}
    \SetAlgoLined
		\vspace{0.5mm}
    \Input{$s_N$: an NDS state; $\mathcal A_G$: goal arrangement
    }
    \Output{$c$: $f(s_N)$}
		\vspace{0.5mm}
		$c\leftarrow g(s_D')$\\
        $c\leftarrow$ Add manipulation costs.\\
        $P\leftarrow$ All deterministic waypoints from $s_D'$ to $s_N$.\\
        $c\leftarrow c+dist(P)$\\
        \For{$o_i$ away from goal in $s_N$}
        {
        \If{$o_i$ in buffer}
        {$c\leftarrow c+$distanceRefinement($s_N$, $o_i$, $\mathcal A_G$)}
        \lElse
        {
        $c\leftarrow c+dist(s_N[o_i],\mathcal A_G[o_i])$
        }
        }
		\Return $c$\\
\end{small}
\end{algorithm}

Algo.~\ref{alg:refine} computes the traveling cost in $f(s_N)$ related to a buffer pose $p_b$ in addition to the straight line path between its neighboring deterministic waypoints (Algo.~\ref{alg:fx} Line 3).
For an object $o_i$ at a buffer, the related traveling cost involves three deterministic points$\{p_s, p_n, p_g\}$: $p_s$ and $p_n$ are the deterministic poses right before and after visiting the buffer.
$p_g$ is the goal pose of $o_i$.
$dist(p_b,p_s)+dist(p_b,p_n)$ is in $g(s_N)$ and $dist(p_b,p_g)$ is in $h(s_N)$.
In the EE scenario, if $\{p_s, p_n, p_g\}$ forms a triangle in x-y space, the distance sum is minimized when $p_b$ is the Fermat point of $\Delta p_sp_np_g$. If $\{p_s, p_n, p_g\}$ forms a line instead, the optimal $p_b$ is at the pose in the middle.
In the MB scenario, we minimize the total base travel. Denote the base positions of $\{p_s, p_n, p_g, p_b\}$ as $\{b_s, b_n, b_g, b_b\}$.
When $b_b$ is not at the three points or their opposite points, there are two points of $\{b_s, b_n, b_g\}$ on one half of the track and another on the other half of the track. 
In the example of Fig.~\ref{fig:workingExample}[Left], if $b_b$ is at the current position, then $b_s$ and $b_n$ are on the left part of the track, and $b_g$ is on the other side.
Moving toward the two-point direction by $d$, $b_b$ can always reduce the total traveling cost by $d$.
Therefore, the extreme points of total distance are $\{b_s, b_n, b_g\}$ and their opposites on the track, and $\{b_s, b_n, b_g\}$ are the minima.
As a result, in MB scenario, the optimal $p_b$ minimizing the total distance to $\{p_s, p_n, p_g\}$ can be chosen among them.

\begin{algorithm}
\begin{small}
    \SetKwInOut{Input}{Input}
    \SetKwInOut{Output}{Output}
    \SetKwComment{Comment}{\% }{}
    \caption{ Distance Refinement}
		\label{alg:refine}
    \SetAlgoLined
		\vspace{0.5mm}
    \Input{$s_N$: an NDS state; $o_i$: object at buffer; $\mathcal A_G$: goal arrangement
    }
    \Output{$c$: additional cost}
		\vspace{0.5mm}
		$p_s\leftarrow$ the pose of $o_i$ before moved to buffer.\\
        $p_n\leftarrow$ the pose that the robot visit after placing $o_i$ to buffer.\\
        $p_g\leftarrow \mathcal A_G[o_i]$.\\ 
        \lIf{EE scenario}
        {
        $p_b\leftarrow$ Fermat point of $\Delta p_sp_np_g$
        }
        \ElseIf{MB scenario}
        {
        $p_b\leftarrow$ The one in $\{p_s,p_n,p_g\}$ with the shortest total distance to the other two.
        }
        \vspace{1mm}
		\Return $\sum_{p\in \{p_s,p_n,p_g\}}(dist(p_b,p))-dist(p_s,p_n)$\\
\end{small}
\end{algorithm}

\subsection{Buffer Allocation}\label{sec:allocation}
We now discuss \ours's buffer allocation process.
Buffer poses are allocated when a new DS is reached, and some object moves to a buffer pose since the last DS node.

\subsubsection{Feasibility of a Buffer Pose}
A buffer pose $p_b$ of $o_i$ is feasible if $o_i$ can be stably placed at $p_b$ without collapsing and $o_i$ should not block other object actions during $o_i$'s stay.
To predict the stability of placement of a general-shaped object, we propose a learning model \model based on Resnest\cite{zhang2022resnest}.
\model consumes two $200*200$ depth images of the surrounding workplace from the top and the placed object from the bottom, respectively.
It outputs the possibility of a successful placement.
The data and labels are generated by PyBullet simulation.
The depth images are synthesized in the planning process based on object poses and the stored point clouds to save time.
In addition to buffer pose stability, we must avoid blocking actions during the object's stay at the buffer pose.
For an action moving an object $o_j$ from $p_j^1$ to $p_j^2$. $o_i$ at the buffer pose needs to avoid $o_j$ at both $p_j^1$ and $p_j^2$.
We add these constraints when we allocate buffer poses.

\begin{figure}[h]
    \centering
    \includegraphics[width=0.8\columnwidth]{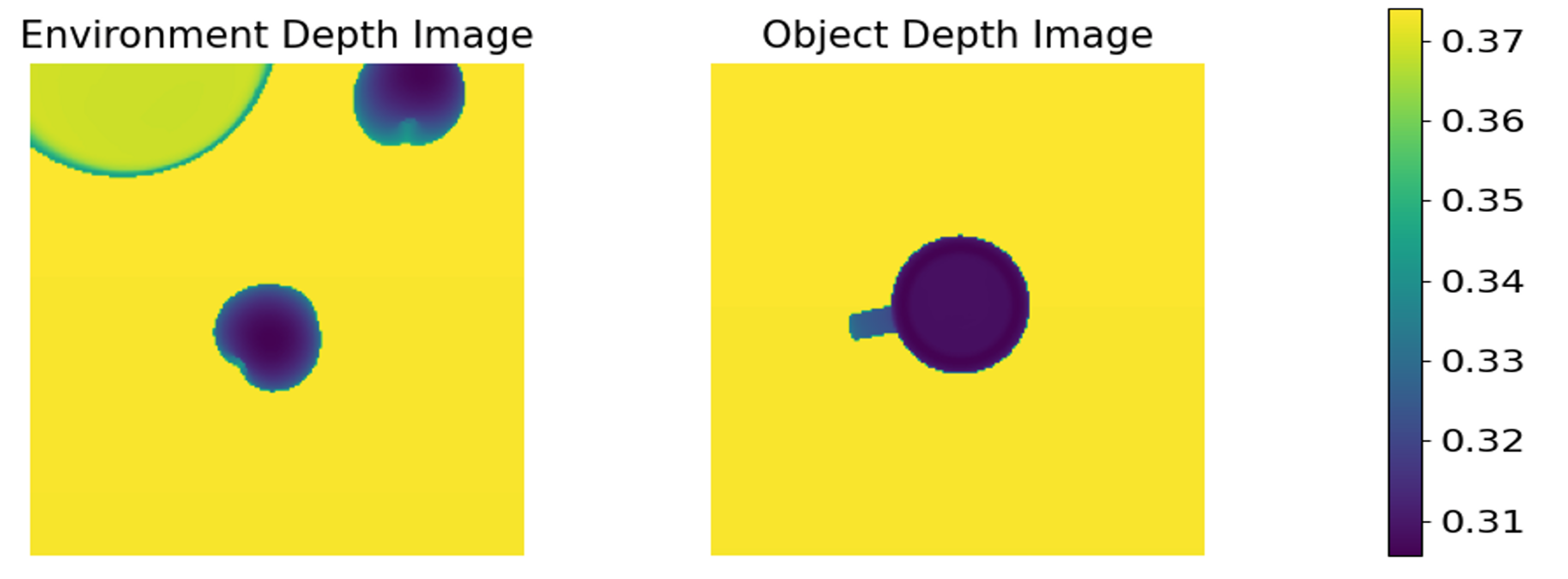}
\vspace{-3mm}
    \caption{An example input of \model when attempting to place a cup right on top of an apple in the environment. The ground truth label given by the simulation is a failure.}
    \label{fig:modelExample}
\end{figure}

\subsubsection{Optimality of a Buffer Pose}
Given the task plan, a set of buffer poses is optimal if the traveling cost is minimized.
For each object needing a buffer pose, we compute $P_b^*$, the set of buffer poses minimizing the traveling cost.
In the traveling cost, the trajectories from the buffer pose $p_b$ to four deterministic poses are involved:
the last deterministic pose the robot visits before placing and picking at $p_b$, 
and the first pose the robot visits after placing and picking at $p_b$.

In the EE scenario, if the four points form a quadrilateral, the total distance is minimized when $p_b$ is placed at the intersection of the diagonal lines. Otherwise, if two or more points overlap, the optimal $p_b$ is the point among four points minimizing the traveling cost.
Therefore, $P_b^*$ in EE scenario is a set of poses with above-computed $x,y$.

In the MB scenario, let $b_b$ be the mobile base position when visiting $p_b$.
And denote the mobile base positions of the four involved poses as $P=\{b_1,b_2,b_3,b_4\}$.
The four points and their opposites $\widehat{P}=\{\widehat{b_1},\widehat{b_2},\widehat{b_3},\widehat{b_4}\}$ partition the track into up to eight segments.
Similar to the case in Algo.~\ref{alg:refine},
$\widehat{P}\bigcup P$ are extreme points of the distance sum.
And the points with the minimum value are optimal $b_b$ solutions.
Moreover, for any of the eight segments, if both of its endpoints are with the minimum value, the whole segments are optimal $b_b$ solutions.
Therefore, in MB scenario, $P_b^*$ is a set of poses whose corresponding mobile base positions are at the points or segments computed above.

\subsubsection{Buffer Sampling}
Algo.~\ref{alg:sampling} handles buffer sampling.
When multiple objects need buffers since the last DS, we sample buffers for them one after another (Line 1-3).
The sampling order is sorted by the time the object is placed in the buffer.
In Lines 4-5, we collect information for pose feasibility checks.
$E$ is used to predict the stability of a buffer pose.
$A$ contains object footprints that $o_i$ must avoid during the buffer stay.
We first sample buffer poses in $P_b^*$ (Line 8-12), and gradually expand the sampling region when we fail to find a feasible buffer (Line 14).
If we cannot find a feasible buffer when the sampling region covers the tabletop area, we return the latest $P$.
In the case of a failure, we remove $S_D$ from \astar search tree and create another new DS node $S_D''$ at the failing step.
$S_D''$ represents the state before the failing $o_i$ is moved to buffer.
At $S_D''$, all objects at buffers have found feasible buffer poses in the returned $P$.
Therefore, all objects in $S_D''$ are at deterministic poses.
Note that in our implementation, $P_b^*$ in MB scenario is represented by a segment of the mobile base track.
As a result, for Line 9 in MB scenario, we first sample a mobile base position on the track and then sample a pose based on that.

\begin{algorithm}[h]
\begin{small}
    \SetKwInOut{Input}{Input}
    \SetKwInOut{Output}{Output}
    \SetKwComment{Comment}{\% }{}
    \caption{Buffer Sampling}
		\label{alg:sampling}
    \SetAlgoLined
		\vspace{0.5mm}
    \Input{$s_D$: an DS state,\\ 
    }
    \Output{$P$: buffer poses}
		\vspace{0.5mm}
        $B \gets$ Objects go to buffers since the last DS.\\
        $P \gets \{o_i:\emptyset \ \forall o_i \in B\}$\\
        \For{$o_i \in B$}
        {
        $E\leftarrow$ The environment when the buffer is placed.\\
        $A\leftarrow$ A list of poses to avoid based on the task plan.\\
        $P_b^*\leftarrow$ The placing region minimizing traveling cost.\\ 
        \While{$o_i$'s Buffer Not Found}
        {
        \For{$i \gets 0$ \textbf{to} $k-1$}
        {$p_b\leftarrow$ Sample a pose in $P_b^*$.\\
        \If{poseFeasible($p_b,o_i,E,A$)}
        {
        Add $p_b$ to $P$;\\
        \textbf{go to} Line 3;
        }
        }
        \lIf{$P_b^*$ is the whole tabletop region}
        {\Return $P$}
        \lElse{
        $P_b^* \gets$ expand($P_b^*$)}
        }
        }
        \Return $P$
\end{small}
\end{algorithm}

\subsection{Optimality of \ours}
Our \ours consists of two main components: high-level \astar search and low-level buffer allocation process.
In terms of the high-level lazy \astar,
the $h(s)$ of DS is consistent, and $g(s)$ of DS does not rely on $g(s)$ of NDSs.
Therefore, by only considering DSs, lazy \astar is a standard \astar with global optimality.
Additionally, $f(s)$ values for NDSs underestimate the actual cost.
Therefore, high-level lazy \astar is globally optimal in its domain.
Note that the formulated buffer allocation problem is an established \emph{Constraints Optimization Problem}, which can be solved optimally if we assume \model makes correct stability predictions.
By replacing the buffer sampling with an optimal solver, \ours is globally optimal.

\section{Evaluation}\label{sec:experiments}
We present simulation evaluations on \ours and compare them with state-of-the-art rearrangement planners implemented in Python.
For each experiment, costs and computation time are averages of test cases finished by all methods.
Experiments are executed on an Intel$^\circledR$ Xeon$^\circledR$ CPU at 3.00GHz.
To measure instance difficulty, we define the density level of workspace $\rho$ as $\dfrac{\sum_{o\in \mathcal O} S(o)}{S_T}$, where $S(o)$ is the footprint size of the object $o$ and $S_T$ is the size of the tabletop region.
We set $C=10$ in cost function $\mathcal J(\Pi)$.

\begin{figure}[h]
    \centering
    \includegraphics[width=0.35\textwidth]{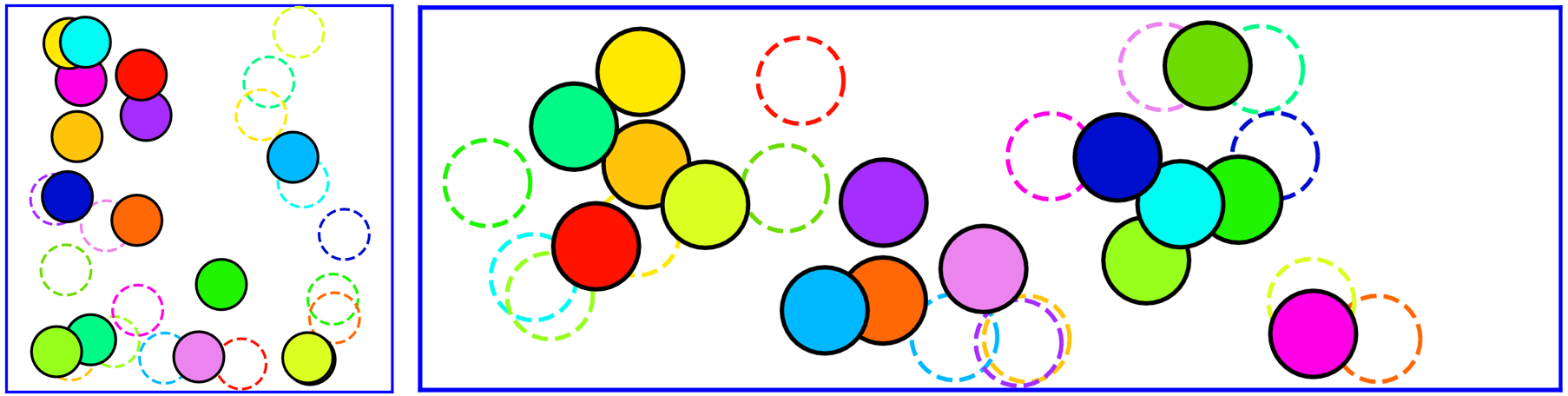}
    \includegraphics[width=0.09\textwidth]{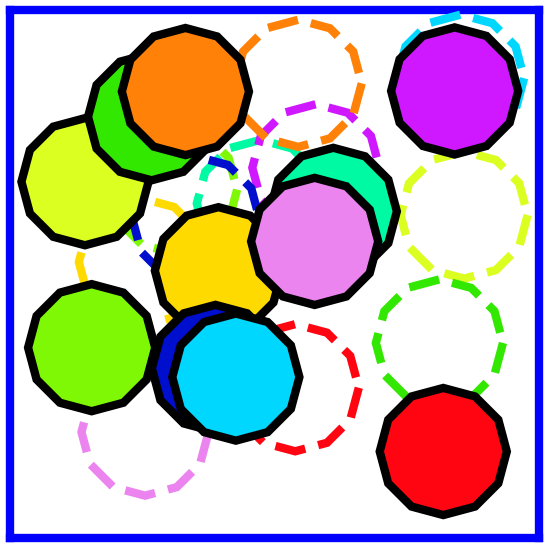}
    \caption{Examples of disc instances. [Left] EE scenario with $\rho=0.2$; [Middle] MB scenario with $\rho=0.2$; [Right] EE scenario with $\rho=0.5$. Colored and transparent discs represent the initial and goal arrangements respectively. }
    \label{fig:discInstance}
\end{figure}

\begin{figure}[h]
\vspace{-3mm}
    \centering
    \includegraphics[width=0.48\textwidth]{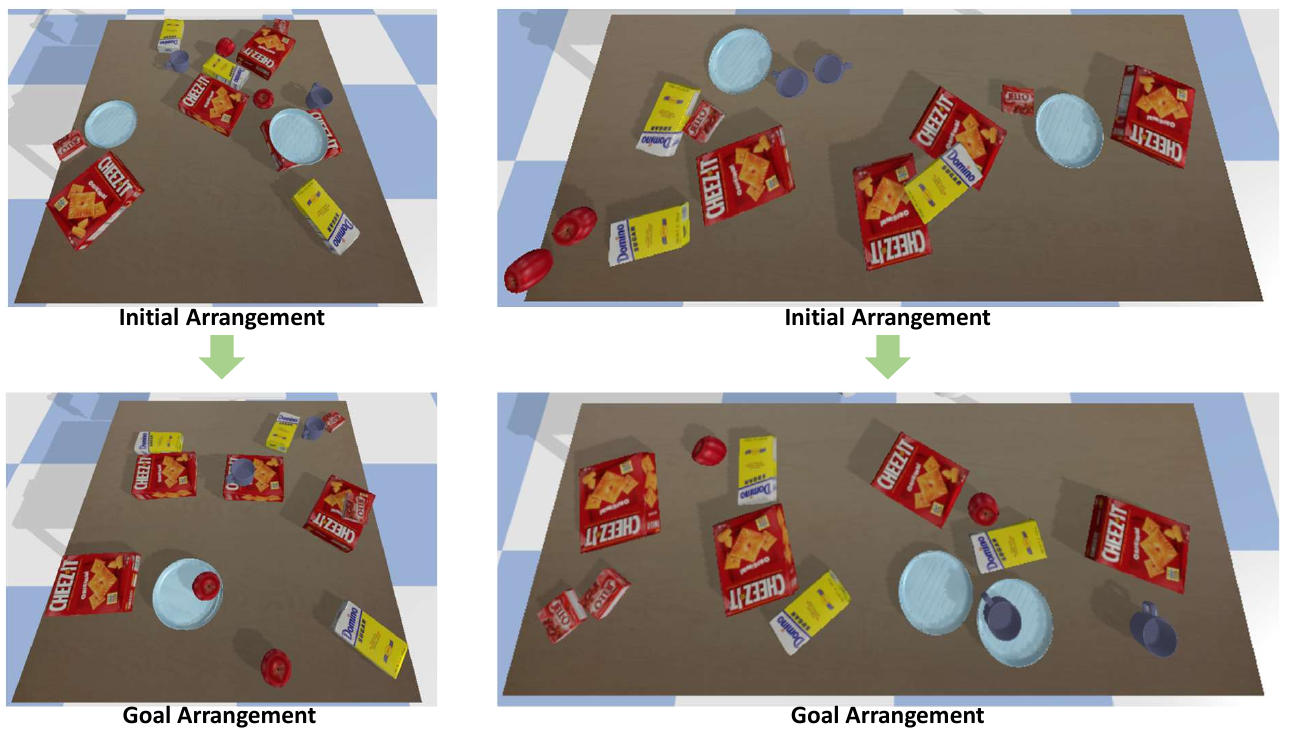}
    \vspace{-4mm}
    \caption{Examples of instances with general-shaped objects in [Left] EE and [Right] MB scenarios.}
    \label{fig:ycbInstance}
\end{figure}

We first compare \ours-Full, \ours-Action, Greedy-Sampling
, MCTS\cite{labbe2020monte}, and TRLB\cite{gao2022fast} 
in disc instances (Fig.~\ref{fig:discInstance}) without \model.
\ours-Full is our \ours minimizing $\mathcal J(\Pi)$.
\ours-Action only considers manipulation costs, i.e. minimizes the number of pick-n-places.
Greedy-Sampling maintains an A$^*$ search tree with $\mathcal J(\Pi)$. Instead of lazy buffer allocation, it samples buffer poses immediately as close to goal poses as possible when objects are moved to buffers.
MCTS and TRLB compute rearrangement plans using Monte-Carlo tree search with random buffer sampling and bi-directional tree search with lazy buffer allocation respectively.
These two planners only consider buffer poses in the free space.
In disc instances, we assume all poses are stable to be placed on but the robot cannot manipulate an object if another object is on top.
The table sizes of EE and MB instances are $1m\times 1m$ and $3m\times 1m$, respectively.
In disc instances with different numbers of objects, we adjust the disc radius to keep $\rho$ constant.
We also test \ours-Full and \ours-Action in general-shaped object instances (Fig.~\ref{fig:ycbInstance}) with \model for buffer pose sampling.

\subsection{Disc Rearrangement in Simulation}
\begin{figure}[ht]
    \centering
    \includegraphics[width=0.48\textwidth]{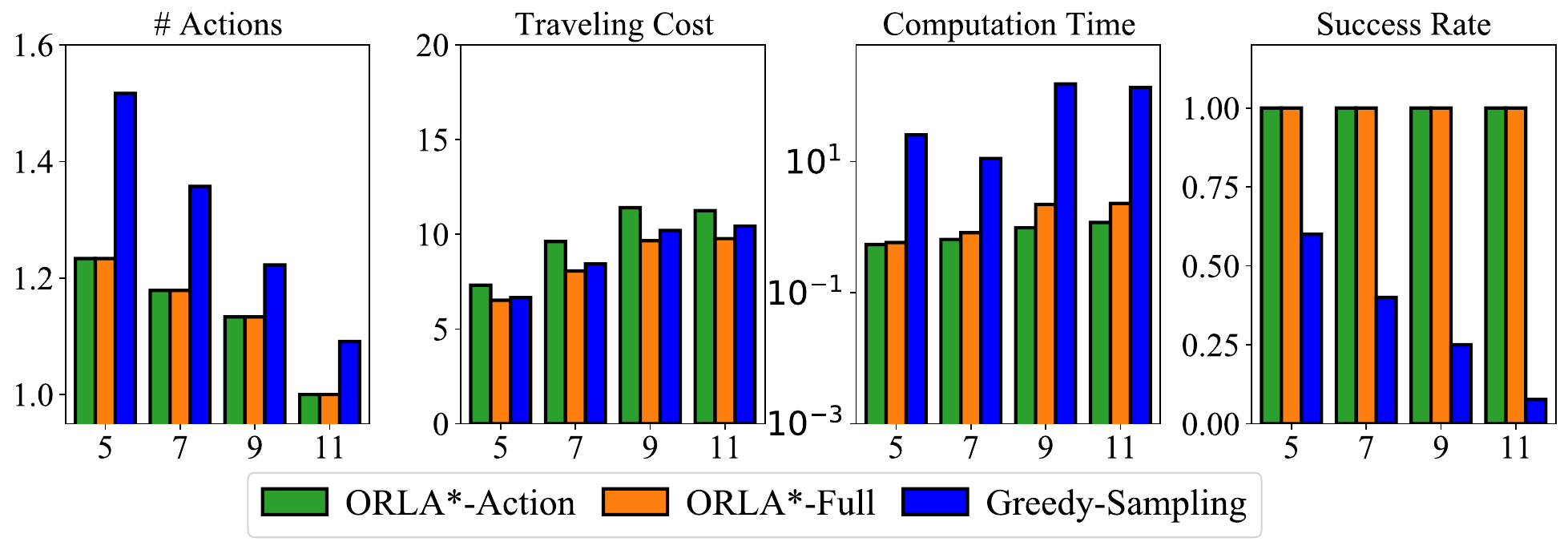}
    \caption{Algorithm performance in EE disc instances with $\rho=0.2$ and 5-11 objects. (a) $\#$ pick-n-places in solutions as multiplies of $|\mathcal O|$. (b) Traveling cost (m). (c) Computation time (secs). (d) Success rate. }
    \label{fig:discs11}
\end{figure}

Fig.~\ref{fig:discs11} shows algorithm performance in EE scenario with $\rho=0.2$. 
Comparing \ours-Full and \ours-Action, \ours-Full saves around $15\%$ path length without an increase in the manipulation cost.
Without the lazy buffer strategy, Greedy-Sampling spends more time on planning but yields much worse plans in general, e.g., additional $23\%$ actions in 5-discs instances.
That is because \ours-Full allocates buffer poses avoiding future actions while immediately sampled buffer poses in Greedy-Sampling may block some of these actions. 
Due to the repeated buffer sampling, Greedy-Sampling can only solve $40\%$ instances in $7-$object instances.
However, Greedy-Sampling has reasonably good performance in the total path length despite the large number of actions, so allocating buffers near the goal pose is a good strategy for saving traveling costs.
We also note that the number of actions as multiplies of $|\mathcal O|$ reduces as $|\mathcal O|$ increases, which indicates a reducing difficulty in rearrangement.
That is because given a fixed density level, as $|\mathcal O|$ increases, the relative size of each object to the workspace is smaller, which makes it easier to find valid buffer poses.

\begin{figure}[ht]
    \centering
    \includegraphics[width=0.5\textwidth]{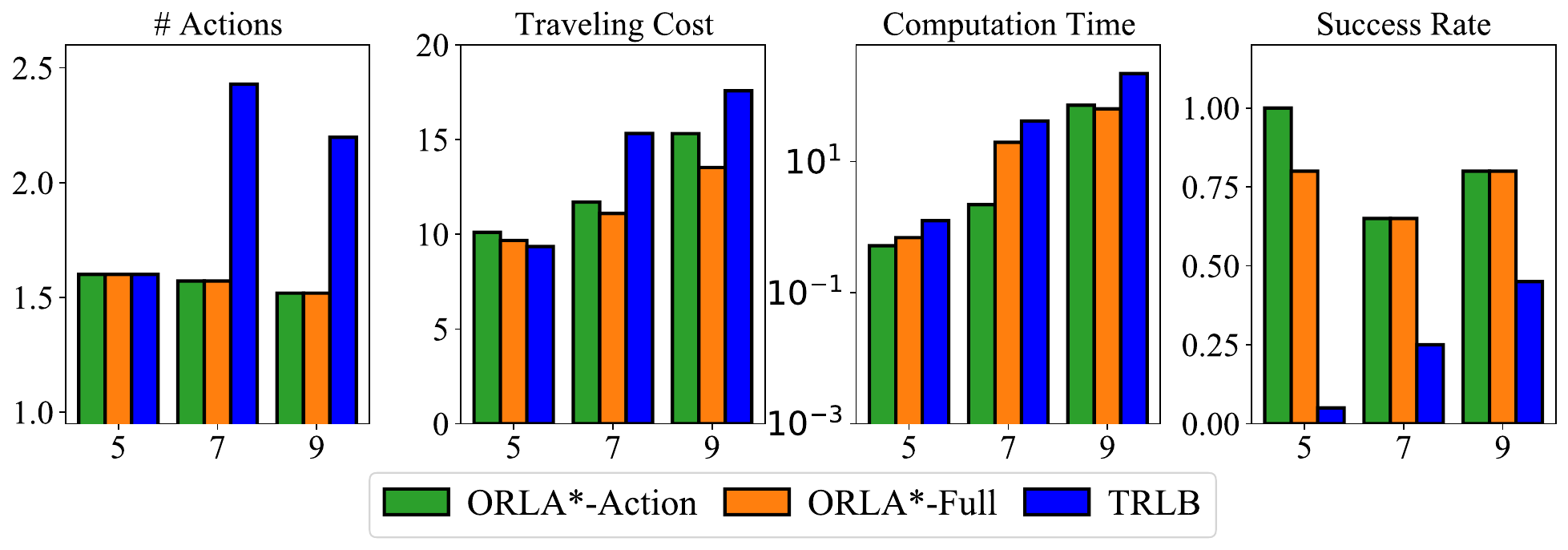}
    \caption{Algorithm performance in EE disc instances with $\rho=0.5$ and 5-11 objects. (a) $\#$ pick-n-places in solutions as multiplies of $|\mathcal O|$. (b) Traveling cost (m). (c) Computation time (secs). (d) Success rate. }
    \label{fig:disc_11_dense}
\end{figure}

We also compare \ours variants with TRLB and MCTS in dense disc instances ($\rho=0.5$) of EE scenario. 
An example of the dense instance is shown in Fig.~\ref{fig:discInstance}[Right]. 
While MCTS fails in all test cases, the results of other methods are shown in Fig.~\ref{fig:disc_11_dense}.
Comparing \ours-Action and TRLB, the results suggest that considering buffer poses on top of other objects not only effectively increases the success rate, but also provides shorter paths with lower traveling cost and manipulation cost in dense test cases.
Comparing \ours variants, \ours-Full saves $4.2\%-11.7\%$ traveling cost, but \ours-Action has a much higher success rate in 5-object instances, which is the hardest for buffer sampling as mentioned previously.

\begin{figure}[ht]
    \centering
    \includegraphics[width=0.48\textwidth]{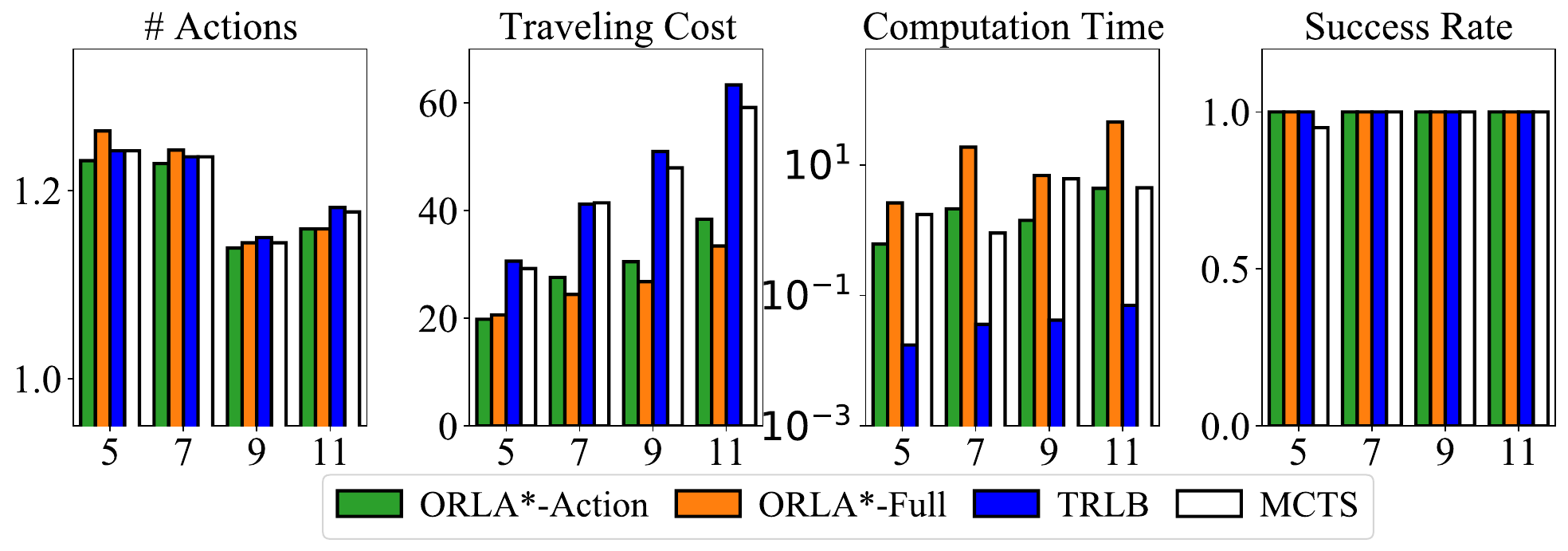}
    \caption{Algorithm performance in MB disc instances with $\rho=0.2$ and 5-11 objects. (a) $\#$ pick-n-places in solutions as multiplies of $|\mathcal O|$. (b) Traveling cost (m). (c) Computation time (secs). (d) Success rate. }
    \label{fig:discs31}
\end{figure}

Fig.~\ref{fig:discs31} shows algorithm performance in MB scenario with $\rho=0.2$.
Compared with TRLB and MCTS, \ours variants save $32.2\%-47.4\%$ traveling cost, which indicates the performance gain of temporary object placement on top of other objects in MB scenarios.
Compared with \ours-Action, \ours-Full saves $13.0\%$ traveling costs on average in 11-disc instances, but spends more time in computation.

To summarize, in disc rearrangement experiments, we have the following conclusions:
First, intelligent temporary object placement on top of other objects with \ours computes low-cost plans in dense test cases and reduces traveling costs in MB scenario.
Second, considering traveling costs in the \ours framework effectively reduces traveling costs without an increase in manipulation costs but induces additional computation time.
Finally, the results suggest that lazy buffer allocation improves solution quality and saves computation time.

\subsection{Qualitative Analysis of \model}
\begin{figure}[ht]
    \centering
    \includegraphics[width=0.48\textwidth]{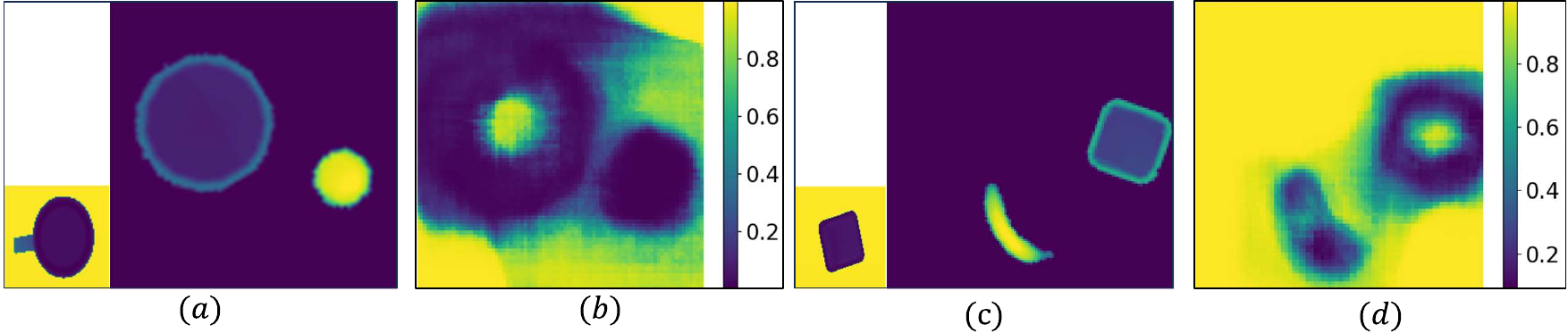}
    \vspace{-5mm}
    \caption{\textbf{(a)} A synthesized environment height map and the depth image of the placed object from the bottom. All objects are from the training set. \textbf{(b)} The corresponding stability prediction distribution of (a). \textbf{(c)} A synthesized environment height map and the depth image of the placed object from the bottom. All objects are outside the training set. \textbf{(d)} The corresponding stability prediction distribution of (c). }
    \label{fig:prediction}
\end{figure}

Regarding the placement stability prediction model \model, we train it with four types of objects in Pybullet simulator: an apple, a pear, a plate, and a cup.
Fig.~\ref{fig:prediction}(a) shows the prediction distribution of a scene with the plate and the apple when placing the cup.
We sample $8$ poses of the placed object with different orientations for each point in the distribution map. 
The distribution shows the average of the $8$ output probabilities.
When the cup pose is sampled far from both objects (the top-right corner and the bottom-left corner) or on the plate, \model supports placements with outputs around $0.999$ and $0.90$, respectively. 
When the cup pose is sampled at the rim of the plate and on top of the apple, \model also rejects placements with outputs around $0.02$ and $0.005$.
Due to the existence of the handle, \model is conservative when the cup is placed close to the plate rim. 
\vspace{-3mm}

To show the model's generalization ability, we present the prediction distribution of a scene with untrained objects in Fig.~\ref{fig:prediction}(c). In this environment, we place a square plate and a banana.
The plate's size and shape differ from those in the training set.
And not to say the banana.
The placed object is a small tea box while we do not have any cuboid object in the training set.
As shown in Fig.~\ref{fig:prediction}(b), \model clearly judges the stability when placing the novel box into the workspace.


\subsection{General-Shaped Object Rearrangement in Simulation}
\begin{figure}[ht]
    \centering
    \includegraphics[width=0.48\textwidth]{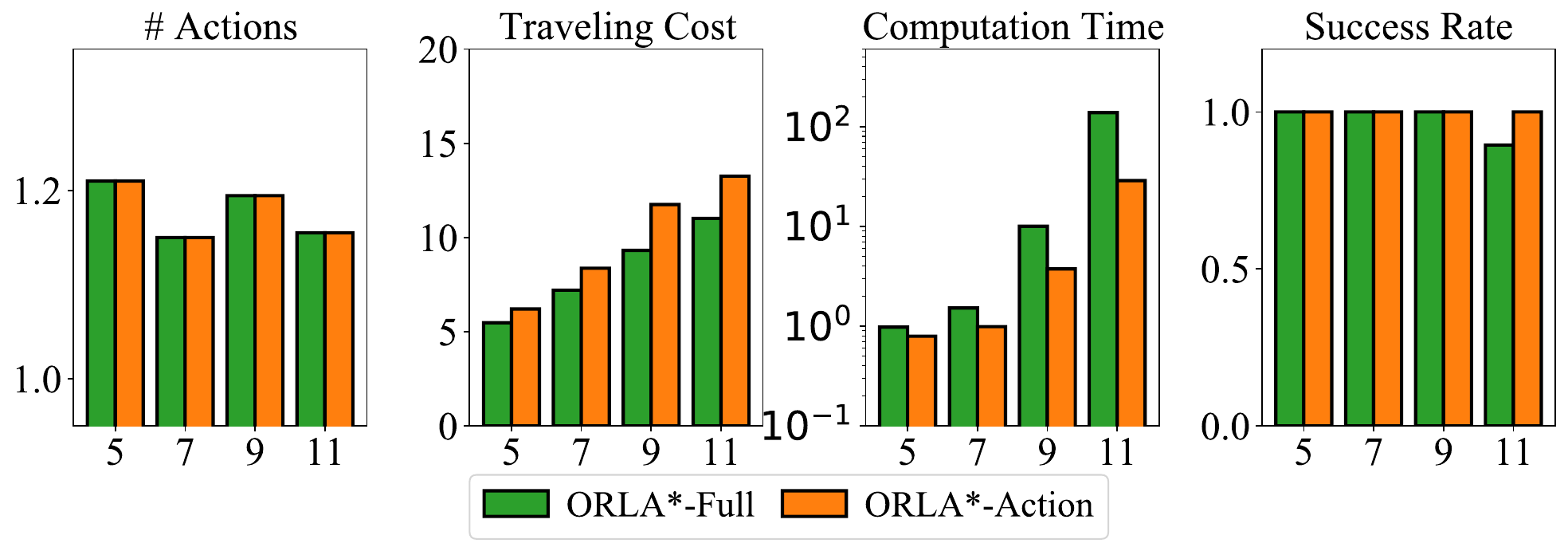}
    \includegraphics[width=0.48\textwidth]{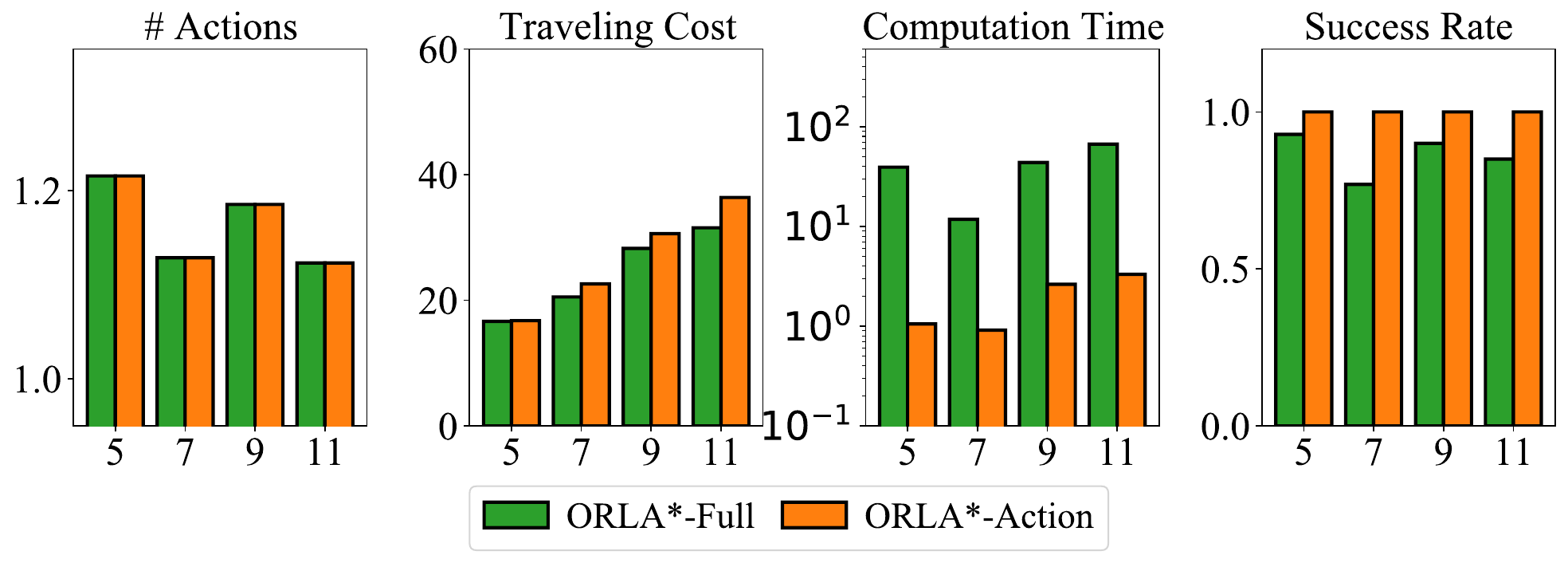}
\vspace{-5mm}
    \caption{Algorithm performance in [Top] EE scenario and [Bottom] MB scenario with general-shaped objects. $\rho=0.25$ and 5-11 objects. (a) $\#$ pick-n-places in solutions as multiplies of $|\mathcal O|$. (b) Traveling cost (m). (c) Computation time (secs). (d) Success rate. }
    \label{fig:ycb11}
\end{figure}

In general-shaped instances, we equip \ours methods with \model.
Given a set of objects in the workspace, we compute the total area of object footprints and adjust the size of the tabletop region to keep $\rho=0.25$.
Fig.~\ref{fig:ycb11} presents algorithm performance in EE and MB scenarios, respectively.
In this scenario, we use \model to decide whether an object is safe to be temporarily placed on top of another.
Due to the inference time for \model in buffer sampling, the computation time of \ours methods is longer than that in disc experiments.
In $11-$object instances, \ours-Full saves $16.7\%$ and $13\%$ traveling cost than \ours-Action in EE and MB scenarios respectively.
However, \ours-Action maintains $100\%$ success rate while \ours-Full fails in $13.5\%$ MB test cases.

\section{Conclusion}\label{sec:conclusion}
We propose \ours, \emph{Object Rearrangement with Lazy A*}, for solving \prob, i.e., layered multi-object rearrangement on large tabletops using a mobile manipulator, which must consider a multitude of factors including: (1) object-object dependencies, (2) jointly optimizing end effector and mobile base travel, and (3) object pile stability. Building on a carefully analysis of  \prob's optimality structure, \ours successfully addresses all the challenges, delivering superior performance on both speed and solution quality, in comparison with the previous solutions applicable to \prob. 
\newpage
\bibliographystyle{format/IEEEtran}
\bibliography{bib/c}

\end{document}